# A Training Method For *VideoPose3D* With Ideology of Action Recognition

Hao Bai
*Zhejiang University – University of Illinois at Urbana-Champaign Institute, ZJUI*
Haining, Jiaxing
Haob.19@intl.zju.edu.cn

*Abstract*—Action recognition and pose estimation from videos are closely related to understand human motions, but more literature focuses on how to solve pose estimation tasks alone from action recognition. This research shows a faster and more flexible training method for *VideoPose3D* which is based on action recognition. This model is fed with the same type of action as the type that will be estimated, and different types of actions can be trained separately. Evidence has shown that, for common pose-estimation tasks, this model requires a relatively small amount of data to carry out similar results with the original research, and for action-oriented tasks, it outperforms the original research by 4.5% with a limited receptive field size and training epoch on Velocity Error of MPJPE. This model can handle both action-oriented and common pose-estimation problems.

## I. Introduction

Video-based human motion analysis attempts to grasp the movements of the human body using computer vision techniques, like pose estimation and motion analysis. Pose estimation is used to rebuild the movements of positions of human body joints inside a sequence of images, and action recognition is a technique to classify different kinds of action typically with a deep learning approach. Lower estimation error in pose estimation and higher classification correctness means better performance.

The effect between pose estimation and action recognition has been proved to be mutual. Pose estimation and action recognition can utilize the same dataset for estimation and training. K. Lee *et al.* proposed an idea of P-LSTM utilizing the dataset *Human3.6M* for both training and pose estimation [1] and Hueihan Jhuang *et al.* promoted insights based on a "systematic performance evaluation using thoroughly-annotated data of human actions", and utilizes *Human3.6M* as a dataset for both pose and action [2]. Pose estimation can be used for action recognition, in which field a lower error leads to a more accurate classification. Yao *et al.* gave a distinctive research on the benefits that pose estimation may give to action recognition [3]. Conversely, action priors can also be used to ameliorate the precision of pose estimation. Yao, Angela *et al.* introduced a model that utilizes a 2D action recognition system as a prior for pose estimation and refinement of the action label [4], Umar Iqbal *et al.* utilized information about actions for developing a better performance on pose estimation in monocular videos [5], and Juergen Gall *et al.* gave a particle-based optimization algorithm that can effectively estimate human pose with the results of a 2D action recognition system as a prior distribution [6]. There have also been works combining these two tasks and perform better in each field. Xiaohan Nie *et al.* developed a model with a hierarchical structure capturing the geometric and appearance variations of pose and lateral connections at adjacent frames capturing the action-specific motion information [7], and Diogo *et al.* created a multitask framework for pose estimation from still images and human action recognition from video sequences [8].

In the field of 3D pose estimation by deep neural network (DNN) approach, there are many successful creatures with various types of training methods, like Maximum-Margin Structured training by Li *et al.* [9], Feedback Loop training by Oberweger *et al.* [10], and also the current state-of-the-art Semi-supervised training by Pavllo *et al.* [11], which we mainly concern to ameliorate in our work.

*VideoPose3D* has been very successful because of its high and estimation precision with datasets *HumanEva* [12] and *Human3.6M*[1] [13]. However, this project takes all types of actions as training data instead of a certain kind, but there have been lots of successful cases when using action recognition for pose estimations works well for improving its performance.

This inspiration gives rise to our research. Concerned about the possible utility that action classification information can give to pose estimation in DNN, our experiments exploit a multi-batch action-sensitive training approach for pose estimation with each action.

## II. Related Works

### A. VideoPose3D

*VideoPose3D* was the state-of-art model which utilizes "a fully convolutional model based on dilated temporal convolutions over 2D key points" [11]. In early researches, the main solution was to use Recurrent Neural Networks (RNN) [14], but the research *VideoPose3D* utilized the convolutional neural network (CNN) to gain an efficient result on the dataset *Human3.6M*, with the inspiration that many authors had mentioned CNN in temporal models.

As a further review of the learning network, *VideoPose3D* takes a temporal sequence of images as the input of the training network as an alternative to RNN, as is shown in

---

[1] *Human3.6M* has 11 subjects and 4 viewpoints in total, and 15 types of actions for each subject.

**Figure 1**. This ideology has been widely accepted in academia, and there have been lots of successful cases utilizing this method, e.g. the spatial-temporal-CNN used for crowd counting in videos [15], and LSTM-CNN used for face anti-spoofing [16]. However, CNN still bears a mathematically natural trait of parallelism and it's time-consuming to use a temporal CNN when the amount of data is huge [17].

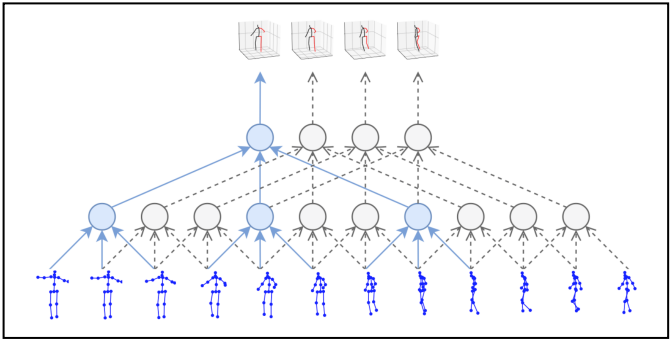

Fig. 1. The temporal convolutional model by [11].

### B. Action Recognition

A remarkable research on applying action recognition on pose estimation was carried by Angela Yao in 2011, shown in **Figure 2** [3]. This model begins with a 2D appearance-based action recognition based on low-level appearance features, after which the outputs of the 2D action recognition are used as a prior distribution for the particle-based optimization for 3D pose estimation. Finally, the 3D pose-based action recognition is performed based on pose features extracted from the estimated poses.

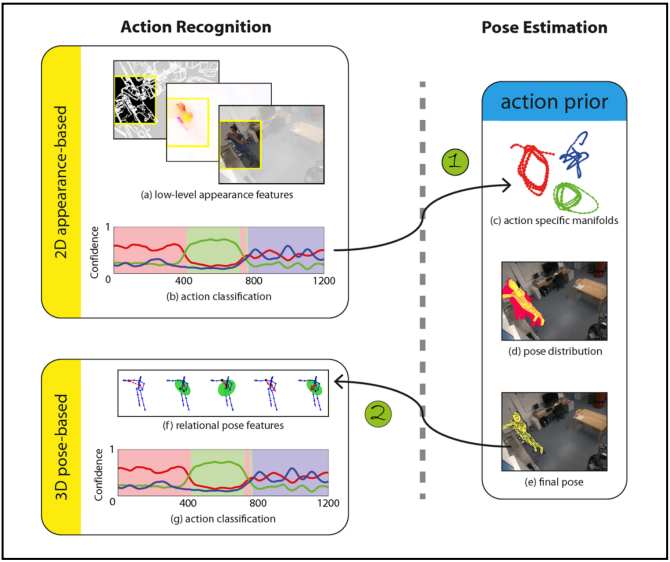

Fig. 2. Coupled Action Recognition and Pose Estimation model by [3].

### III. MODELING

The code of this project is available under the open-source *MIT* License at the GitHub Page[2]. The detailed experimental setup method can be explored in *readme.md* in the GitHub Repository. To reproduce the results, it's preferred to test for multiple times and use *bash* scripts provided in *lsf.sh*. When estimating results, expect an error of 5%.

The action-based training model can be applied in two kinds of problems. The first one is the common pose-estimation problems, just like what *VideoPose3D* aims to do. This kind of task gives all kinds of data and estimates errors for all kinds of actions. The second one is the action-oriented pose estimation problems, which gives all kinds of data but the grading standard mainly focuses on one certain action.

This research focuses on using certain types of actions for training data with the dataset *Human3.6M*, which is the same as *VideoPose3D*.

### A. Model for Common Problems

As mentioned before, *Human3.6M* has 15 actions for each subject, and *VideoPose3D* takes all types of data as training data, as shown in the left part of **Figure 3**, and estimates results for all actions in the end. Our work, instead, takes only one action for one round of training, and estimates the result for this certain action in the estimation period after 15 rounds of training data of the same amount of action, as shown in the right part of **Figure 3**. This process can either be implemented in iterations or parallel.

For this kind of problem, one training epoch in *VideoPose3D* is equivalent to approximate 15 training epochs in our experiment if we utilize a dataset with the number of frames approximately the same among different actions, such as *Human3.6M*, because the training data amount for each action must be the same. If we represent $t_0$ as one original *VideoPose3D* epoch, $t_{unit}$ as one epoch in our work (it'll be called **unit epoch** in later parts of this paper), and $n_{ac}$ as the number of actions in the used dataset, the relationship between them can be expressed by **Formula (1)**.

$$t_0 = t_{unit} \cdot n_{ac} \quad (1)$$

In other cases, when the numbers of frames among different actions are different, it's preferable to use the more common way to calculate the epochs. If we represent $f$ as the function to calculate the total amount of frames of one action and $action$ as the required action (which means $action \in action$) we can calculate the training epochs of this experiment using **Formula (2)**.

$$\frac{t_0}{t_{unit}} = \left\lfloor \frac{\sum_{i \in actions} f(i)}{f(action)} \right\rfloor \quad (2)$$

### B. Model for Action-oriented Problems

The action-oriented tasks make a difference in the data amount of experiments and the result grading standard: the total amount of training data is limited, and the only grading standard is one action. For example, if there is a large dataset with 5000 images for each action, one model can only feed in 6000 images. In this case, *VideoPose3D* takes 400 images for each action, but our model takes 6000 images for exactly the required action. If we represent $n_{VP}$ as the number of data in *VideoPose3D* for a certain action, and $n_{AB}$ as the number of data in our action-based model for a certain action, the relationship between the number of training data can be represented mathematically as **Formula (3)**.

$$\sum_{i \in actions} n_{VP}(i) = n_{AB}(action) \quad (3)$$

---

[2] https://github. com/BiEchi/Pose3dDirectionalTraining

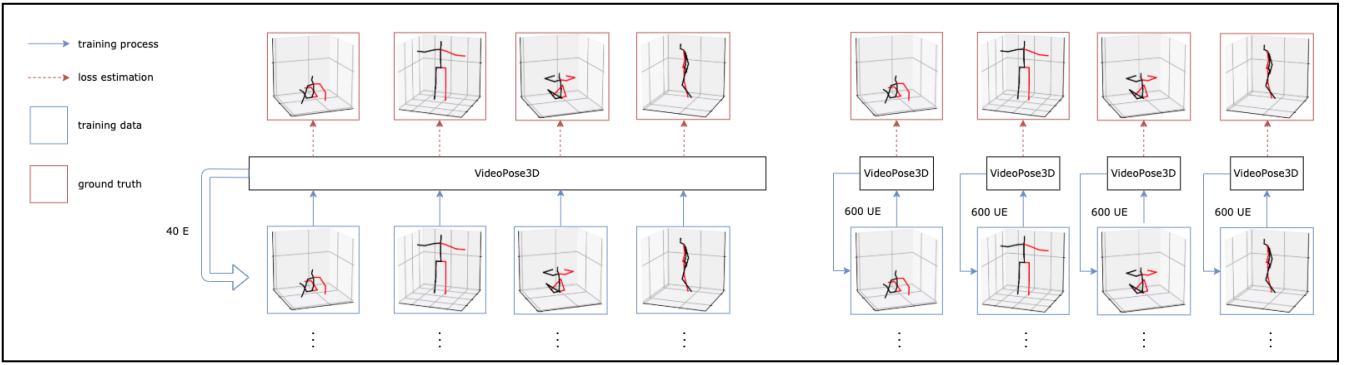

Fig. 3. Comparison of models between our model and *VideoPose3D* for common pose-estimation tasks

TABLE I.　　PROTOCOL I. MPJPE ERROR.

| Arguments | Model | Dir. | Disc. | Eat | Greet | Phone | Photo | Pose | Purch. | Sit | SitD. | Smoke | Wait | WkD. | Walk | WkT. | Avg |
|---|---|---|---|---|---|---|---|---|---|---|---|---|---|---|---|---|---|
| F=1, UE=15 | VideoPose3D | 67.0 | **65.8** | 68.7 | 72.5 | **74.2** | **87.2** | 65.5 | 73.1 | 85.6 | 117 | 73.6 | **70.8** | 81.1 | 63.5 | 67.3 | 75.5 |
|  | Ours | **61.9** | 69.2 | **62.0** | **68.4** | 75.7 | 88.2 | 74.5 | 76.9 | **81.5** | **97.9** | **71.1** | 80.9 | **80.7** | **49.9** | **59.2** | **73.2** |
| F=27, UE=15 | VideoPose3D | **54.5** | **61.4** | 56.3 | **58.6** | **61.3** | **68.6** | **57.6** | **60.6** | 70.5 | 84.7 | 60.5 | **59.1** | 68.2 | 51.8 | **53.1** | **61.8** |
|  | Ours | 70.5 | 71.9 | **52.1** | 61.4 | 69.0 | 82.6 | 70.6 | 82.1 | **70.4** | **79.4** | **60.3** | 79.5 | **56.1** | **44.2** | 56.2 | 67.1 |
| F=243, UE=1200 | Pavlakos [18] | 67.4 | 71.9 | 66.7 | 69.1 | 72.0 | 77.0 | 65.0 | 68.3 | 83.7 | 96.5 | 71.7 | 65.8 | 74.9 | 59.1 | 63.2 | 71.9 |
|  | Luvizon [8] | 49.2 | 51.6 | 47.6 | 50.5 | 51.8 | 60.3 | 48.5 | 51.7 | 61.5 | 70.9 | 53.7 | 48.9 | 57.9 | 44.4 | 48.9 | 53.2 |
|  | VideoPose3D | **46.6** | **47.4** | 45.2 | **46.2** | **49.0** | **56.7** | **46.4** | **47.2** | **59.9** | 68.2 | **48.1** | **46.2** | 49.4 | 32.9 | **34.3** | **48.2** |
|  | Ours | 47.9 | 48.8 | **45.1** | 48.4 | 51.7 | 62.8 | 47.1 | 59.4 | 61.2 | **64.7** | 48.2 | 59.3 | **46.6** | **31.4** | 35.5 | 50.5 |

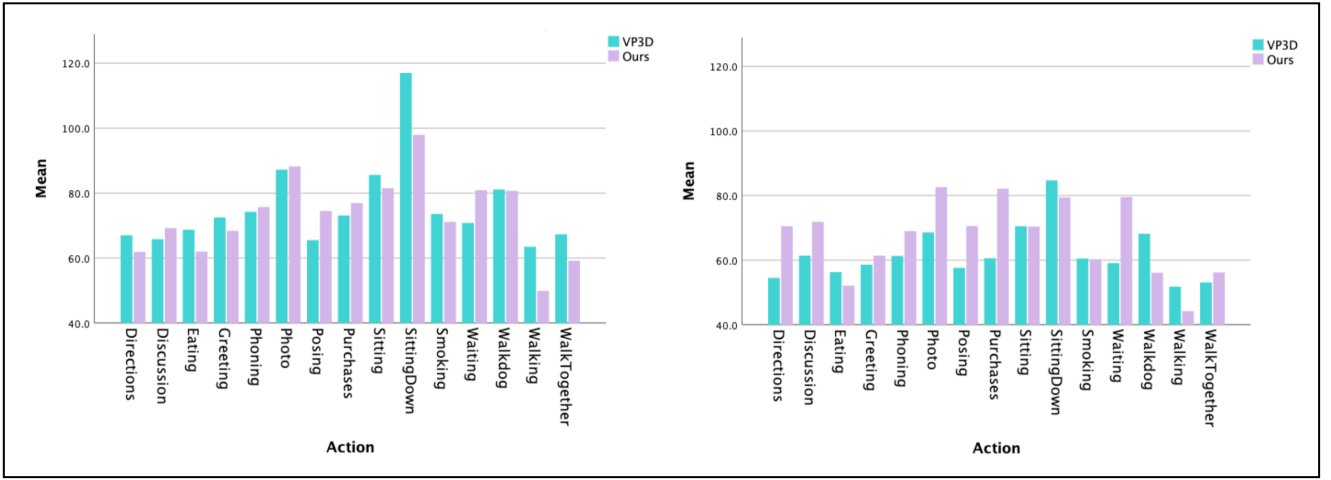

Fig. 4. Visualization for MPJPE error comparison. **Left**: error under 1F, 15UE. **Right**: error under 27F, 15UE.

## IV. SELECTED RESULTS

Although the model for each task is different, the experiments only need to concern about the performance of this work and *VideoPose3D* respectively. We divide the results in two ways, i.e., the results with comparison to training and estimation under different arguments and different time consumption.

### A. Training Under Different Arguments

In this part, we collect and analyze the pose estimation results concerning the two models with different training arguments (i.e., the number of receptive field frames **F**, and the number of Unit Epochs **UE**), 15 different actions, and two different protocols (i.e., MPJPE and Velocity Error of MPJPE).

In the tables, the model with the best performance is marked in bold, and we set three groups of arguments for better comparison.

The direct results for Protocol **MPJPE** are shown in **Table 1**, and the visualization for the error is shown in **Figure 4**. From the results we can draw three main conclusions:

For a common pose-estimation problem, under a limited receptive field size and training epoch, our model outperforms the original research by 3.05%. As mentioned in the third part of this paper, a common pose-estimation problem requires strictly the same amount of training data, which implies that the arguments for both experiments must be the same, so we compare the first row and second row in **Table 1** and the green and pink bar in the left of **Figure 4**.

TABLE II.  PROTOCOL II. V-MPJPE ERROR.

| Arguments | Model | Dir. | Disc. | Eat | Greet | Phone | Photo | Pose | Purch. | Sit | SitD. | Smoke | Wait | WkD. | Walk | WkT. | Avg |
|---|---|---|---|---|---|---|---|---|---|---|---|---|---|---|---|---|---|
| F=1, UE=15 | VideoPose3D | **10.7** | **12.2** | 10.2 | 12.4 | 11.1 | **11.0** | 10.5 | 12.3 | **11.4** | 13.8 | 11.0 | **10.6** | 12.7 | 12.9 | 12.4 | 11.7 |
|  | Ours | 10.9 | 13.0 | **9.59** | **11.8** | **10.3** | 11.4 | **10.1** | **10.5** | 12.2 | 14.6 | **10.9** | 11.5 | **12.5** | **12.6** | **12.2** | **11.6** |
| F=27, UE=15 | VideoPose3D | 3.84 | 4.02 | 3.14 | 4.42 | 3.31 | 3.58 | **3.44** | 3.93 | 3.14 | 4.21 | 3.36 | **3.34** | 4.68 | 4.51 | 4.08 | 3.80 |
|  | Ours | **3.78** | **4.28** | **3.01** | **4.27** | **3.14** | **3.50** | 3.79 | **3.65** | **2.84** | **3.90** | **3.05** | 3.55 | **4.37** | **3.77** | **3.49** | **3.63** |
| F=243, UE=1200 | VideoPose3D | **3.01** | 3.21 | **2.33** | **3.55** | 2.33 | **2.83** | **2.77** | 3.23 | 2.11 | 3.01 | 2.44 | **2.43** | 3.82 | 3.33 | **2.86** | **2.49** |
|  | Ours | 3.14 | **3.20** | 2.48 | 3.62 | **2.31** | 2.91 | 3.04 | **3.11** | **2.07** | **2.88** | **2.23** | 2.73 | **3.66** | **3.04** | 2.97 | 3.08 |

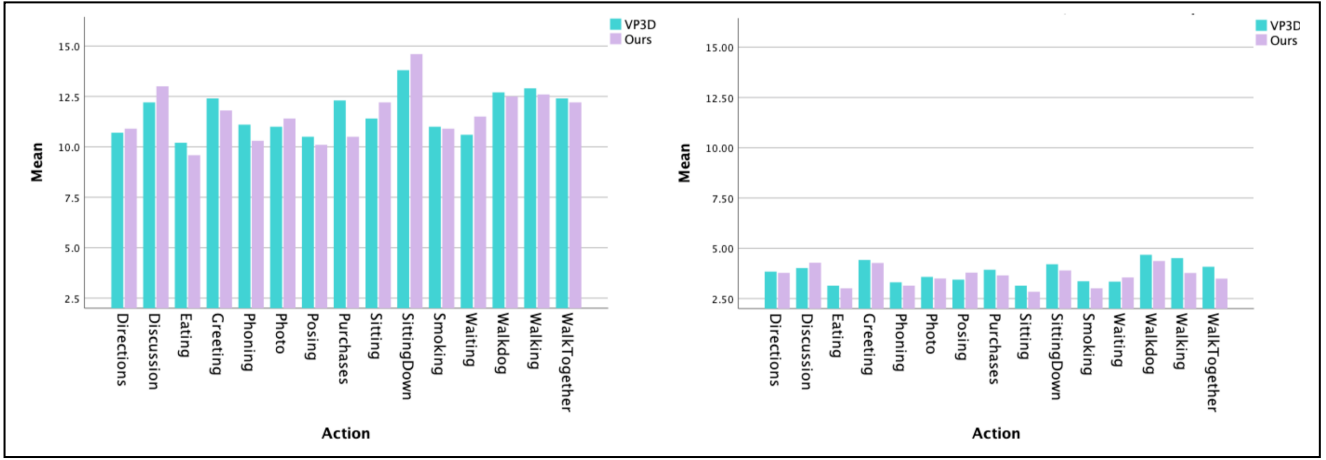

Fig. 5. Visualization for V-MPJPE error comparison. **Left**: error under 1F, 15UE. **Right**: error under 27F, 15UE.

For a common pose-estimation problem, when the receptive field size grows up, the original work behaves a better performance than this work. This observation requires a horizontal comparison between the first two rows and the second two rows. As the field size grows to $3 \cdot 3 \cdot 3$, the errors in the original work was dramatically cut down by 18.2% due to its convolutional network, but our work is only improved by 8.3%.

For a common pose-estimation problem, when a large amount of training resources (data and time) is provided, this model is not as good as the original one, but still better than most other models. This observation is based on the last four rows in **Table 1**. From the table, it's easily seen that the dominant results are still concentrated on the original research, and our model performs better only on a few actions. However, our model still outperforms most other models because it still bears most of the structural advantages from the original work. For example, our model outperforms the model put forward by Pavlakos *et al*. by a great amount of 42.3%, in which experiment the researchers also utilized ground truth bounding boxes for training help [11, 18], and the model by Luvizon *et al*. by 5.3%, which not only made use of ground truth bounding boxes, but also extra data [8, 11].

Besides this protocol, our experiment also covers the **V-MPJPE** protocol, which examines the velocity error difference between the two models. The direct results for them are shown in **Table 2**, and the visualization is shown in **Figure 5**. From the results we can draw one main conclusion:

For a common pose-estimation problem, with limited training resources, our model improves quicker than the original one in velocity error and performs better than the original work overall. According to the first two rows and second two rows, the average velocity error of our model improves 219.6%, which is faster than the original model (207.9%). Moreover, our work outperforms the original work under 27 frames, 15 unit epochs by 4.7%, which indicates a potency of our model to get used to a relatively mild environment.

With all conclusions above, there is one conclusion and one deduction:

In a common pose-estimation task, this work outperforms the original work overall under a limited training resource but performs worse than the original work when the training resource is abundant. As is mentioned in the conclusions above, this model outperforms the original one no matter in MPJPE protocol or V-MPJPE protocol when training resources are strictly limited. As the training resource becomes more abundant, the capability of this model improves slower than the original model overall (excluding V-MPJPE), but still faster than other models. Also, this model improves much faster on V-MPJPE protocol than MPJPE protocol.

A deduction is that, in an action-based pose-estimation task, this work performs better under a limited training resource, and performs better than the original work under an abundant training resource for certain kinds of actions. As we see in **Figure 4** and **Figure 5**, the error difference between this work and the original work is tiny. This work is not as good as the original work under a long time of training mainly because it lacks a variety of data, as this work utilizes the same model as the original one, and the original model was driven by a large variety of data. In an action-oriented pose-estimation task, the variety of this model will be greatly ameliorated, which implies a great improvement of performance. Also, even using a small amount of data, our model still outperforms the original model for certain actions (e.g. discussion and sitting down).

## B. Temporal Comparison in Training Process

The time for training data is a crucial standard for deep learning. One example is the great success of the invention of Faster R-CNN [19]. In this part, we take a deep insight into the time consumption and precision-time rate between the original model and our Eating-based model in **Table 3**, and the temporal relationship between our work and the original work using protocol MPJPE with the lapse of epochs in **Table 4**.

In **Table 3**, $\epsilon_0$ stands for the regulated error when training to the half of the process (in our case 40, as the total epoch is 80), $\epsilon_t$ represents the error after the total epoch of training, and TPR is the abbreviation for the time-precision rate. If we represent TPR as $\theta$, *VideoPose3D* and our model as $i \in \{1, 2\}$, the real time consumed by each training, and the augmenting constant $k = 1.2$, then the calculation of TPR can be represented mathematically by **Formula (4, 5)**.

$$\epsilon_0^{(1,2)} = \frac{\epsilon_{\lfloor\frac{t}{2}\rfloor}^{(1)} + \epsilon_{\lfloor\frac{t}{2}\rfloor}^{(2)}}{2} \cdot k \quad (4)$$

$$\theta^{(i)} = \frac{\epsilon_0^{(1,2)} - \epsilon_t^{(i)}}{t} \quad (5)$$

TABLE III. PROPERTIES COMPARISON WITH F=243, UE=1200.

| Object | VP3D-Avg | Ours-Avg |
|---|---|---|
| TC (40 E) | 84972 sec. | 28848 sec. |
| MPJPE | 50.27 mm | 50.49 mm |
| Velocity-M | 2.66 mm | 3.02 mm |
| TC (80 E) | 176976 sec. | 56921 sec. |
| MPJPE | 48.22 mm | 50.54 mm |
| Velocity-M | 2.49 mm | 3.08 mm |
| MPJPE $\epsilon_0$ | 60.46 | |
| Velo-M $\epsilon_0$ | 3.408 | |
| MPJPE TPR | $6.92 \cdot 10^{-5}$ | $17.4 \cdot 10^{-5}$ |
| Velo-M TPR | $5.19 \cdot 10^{-6}$ | $5.76 \cdot 10^{-6}$ |

From **Table 3** we can easily figure out 2 main conclusions:

For a common pose-estimation problem, our model takes a much shorter time of training but can still carry out similar effects as the original model. According to the first three rows and seconds three rows, we observe a decrease of training cost (TC) by approximately 194.6%, with almost the same MPJPE error and a worse estimation result by about 13.5%.

Our model is proved to have a higher efficiency according to the TPR Protocol. According to **Table 3**, our model has a higher time-precision rate than the original work by 151.4% for MPJPE Protocol and 11.0% for V-MPJPE Protocol.

In **Table 4**, we represent the time-related error results among the average error and error of one specific action for both models. The visualization of this result is shown in **Figure 6**, where our models are colored dark, and the original models are colored light. With these data, we can figure out two more conclusions.

Our model converges faster than the original model. According to **Figure 6**, our model begins to converge at Epoch 6 on average, and at Epoch 4 for eating. In contrast, the original work starts converging around Epoch 50 on average, and cannot converge for eating.

Our model has a more stable training effect on average. From the same figure, we can see the green line (average error for *VideoPose3D*) has a much larger variance than the dark blue line (average error for this work), especially at Epoch 3.

TABLE IV. TEMPORAL COMPARISON WITH F=243, UE=1200

| Epochs | VP3D-Avg | Ours-Avg | VP3D-Eat | Ours-Eat |
|---|---|---|---|---|
| 1 | 59.11 | 57.52 | 54.22 | 51.14 |
| 2 | 56.26 | 52.06 | 52.21 | 48.72 |
| 3 | 61.73 | 51.28 | 51.17 | 47.14 |
| 4 | 54.24 | 51.41 | 49.22 | 45.28 |
| 5 | 51.17 | 51.11 | 49.74 | 45.17 |
| 6 | 50.16 | 50.11 | 51.12 | 46.28 |
| 7 | 50.28 | 50.37 | 50.71 | 45.82 |
| 8 | 50.19 | 50.25 | 50.23 | 47.14 |
| 9 | 50.24 | 50.34 | 49.17 | 47.22 |
| 10 | 50.14 | 50.94 | 49.82 | 46.32 |
| 30 | 49.56 | 50.56 | 47.71 | 45.64 |
| 40 | 50.27 | 50.49 | 46.14 | 45.18 |
| 50 | 48.17 | 50.29 | 45.61 | 45.12 |
| 80 | 48.22 | 50.54 | 45.22 | 45.14 |

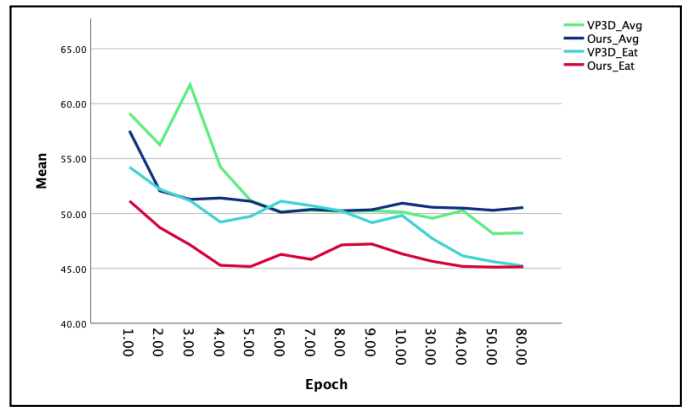

Fig. 6. Visualization of the temporal comparison with F=243, UE=1200.

Wrapping up all the temporal-related conclusions in this part, we can figure out one more deduction:

Our model not only improves the training efficiency of the original work but also decreases the number of epochs for training to the saturation state.[3] The improvement of the training efficiency can be concluded from the performances on TPR Protocol, and the decrement of the number of epochs is observed from a quicker convergence in our model.

## V. CONCLUSIONS AND FUTURE WORK

To keep consistent with the third part of this paper, the conclusion of our work is divided into the meanings for common pose-estimation tasks and action-based pose-estimation tasks.

For common pose-estimation tasks, when the training resource is sufficient, our model provides a quicker and more efficient solution, although the result has a tiny fallback than the original work; when the training is under a strict environment, our model provides a solution with better performance.

---

[3] Note that decreasing the number of epochs is different from decreasing the training time, which depends not only on the number of epochs but also the training time for each epoch.

For action-based pose-estimation tasks, we conjecture, therefore, that whether the training resource is deficient or not, our model can always provide a more efficient and reliable solution.

Future work of this research should contain proof to our conjecture that our model performs better than the original work in action-based pose-estimation tasks.

ACKNOWLEDGMENT

The work was supported by professor Gaoang Wang, who gave rise to the idea of "training from scratch using action-based data". This work would not be completed without encouragement from him.